\documentclass[runningheads]{llncs}

\usepackage{eccv}

\usepackage{eccvabbrv}

\usepackage[table,xcdraw]{xcolor} % 引入 xcolor 包来支持表格背景颜色
\usepackage{graphicx}
\usepackage{booktabs}

\usepackage{makecell}
\usepackage[accsupp]{axessibility}  % Improves PDF readability for those with disabilities.

\usepackage{hyperref}
\usepackage{wrapfig}
\usepackage{orcidlink}

\usepackage{multirow}  % 提供 \multirow

\begin{document}

\title{Next-Frame Decoding for Ultra-Low-Bitrate Image Compression with Video Diffusion Priors} 

\titlerunning{NeFIC}


\author{Yunuo Chen\inst{1} \and
Chuqin Zhou\inst{1} \and
Jiangchuan Li\inst{1} \and
Xiaoyue Ling\inst{1} \and
Bing He\inst{1} \and
Jincheng Dai\inst{2} \and
Li Song\inst{1} \and
Guo Lu\inst{1}\thanks{Corresponding author. E-mail: \texttt{luguo2014@sjtu.edu.cn}.}}

\authorrunning{Y.~Chen et al.}

\institute{Shanghai Jiao Tong University, Shanghai, China \and
Beijing University of Posts and Telecommunications, Beijing, China}

\maketitle

\begin{center}
  \includegraphics[width=\linewidth]{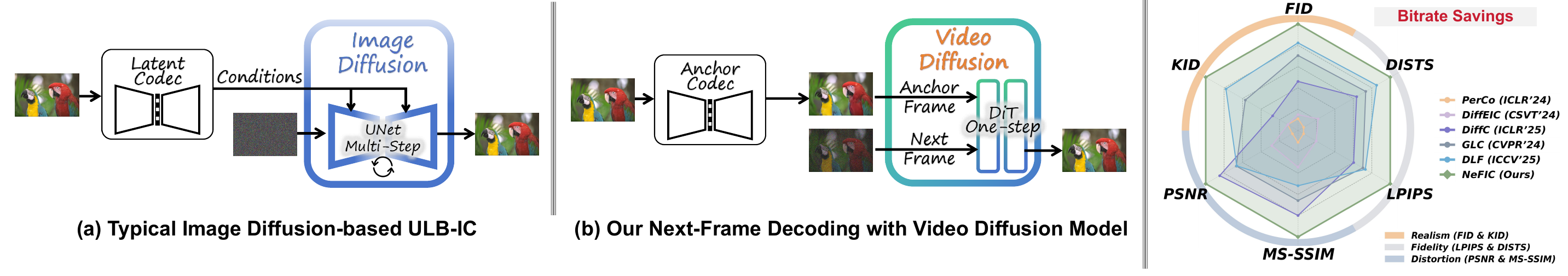}
  \vspace{-15pt}
  \captionof{figure}{(a) Some previous diffusion-based ULB-IC methods transmit latents that serve as conditions for image diffusion models during generative decoding. (b) Our method decodes a compact anchor frame and uses a video diffusion model to temporally evolve this anchor into the target image via next-frame prediction. Radar plot compares our model with existing generative codecs on the CLIC2020 test set.}
  \vspace{-5pt}
  \label{fig:teaser}
\end{center}

\begin{abstract}
We present a novel paradigm for ultra-low-bitrate image compression (ULB-IC) that exploits the ``temporal'' evolution in generative image compression.
Specifically, we define an explicit intermediate state during decoding: a compact anchor frame, which preserves the scene geometry and semantic layout while discarding high-frequency details. We then reinterpret generative decoding as a virtual temporal transition from this anchor to the final reconstructed image.
To model this progression, we leverage a pretrained video diffusion model (VDM) as a temporal prior: the anchor frame serves as the initial frame and the original image as the target frame, transforming the decoding process into a next-frame prediction task.
In contrast to image diffusion-based ULB-IC models, our decoding proceeds from a visible, semantically faithful anchor, which improves both fidelity and realism for perceptual image compression.
Extensive experiments demonstrate that our method achieves superior rate-distortion performance.
On the CLIC2020 test set, our method achieves over \textbf{50\% bitrate savings} across LPIPS, DISTS, FID, and KID compared to DiffC, while also delivering a significant decoding speedup of up to $\times$5.
Code will be released at \url{https://github.com/UnoC-727/NeFIC}.
\keywords{Image Compression \and Generative Compression}
\end{abstract}

% \vspace{-3pt}
\section{Introduction}
\label{sec:intro}
\vspace{-5pt}
The explosive growth of image data has sharply increased storage and transmission burdens, motivating steady progress in high-performance lossy image compression.
Learned image compression (LIC) has advanced rapidly under variational autoencoders (VAEs) and end-to-end optimization \cite{balle2016end, balle2018variational}. Early models relied on CNN-based analysis-synthesis transforms \cite{minnen2018joint, he2022elic, aaai_reference}, while later Transformer- and Mamba-based architectures further improved the rate-distortion (RD) performance \cite{zou2022devil, zhu2022transformer, liu2023learned, scaleup, qin2024mambavc, li2023frequency, zeng2025mambaic, iccv25HPCM, cvpr2025diction, chen2026contentaware_cmic, chen2025s2cformerreorientinglearnedimage, chen2025knowledgeKDIC,du2025largeLLM}. 
Concurrently, the rise of Generative Adversarial Networks (GANs) \cite{GAN} and diffusion models \cite{ddim, DDPM} has reframed ultra-low-bitrate image compression as a pivotal frontier. By leveraging powerful generative priors, diffusion-based codecs can deliver striking perceptual realism at extremely low bitrates. Most existing approaches \cite{aigc_corrdiff, aigc_perco, aigc_diffeic} control the multi-step denoising process via control networks (Fig.\ref{fig:teaser} (a)). However, the unclear influence of the conditional signals on the diffusion process, and the uncertainty of standard Gaussian initialization, together potentially induce semantic drift and weaken faithfulness to the source \cite{chen2024faithdiff, aigc_corrdiff, aigc_resulic, aigc_addingdl}. In addition, the iterative denoising process incurs substantially higher computational cost and decoding latency than conventional VAE-style codecs.
So far, the central challenge for ULB-IC is to achieve simultaneously: (i) strong perceptual realism, (ii) source-content faithfulness, and (iii) practical inference efficiency.

\begin{figure*}[t]
    \centering
\captionsetup{font=footnotesize}
    \includegraphics[width=1\linewidth]{figs/selected_no_neg_13frames_each_vertical_montage.jpg}
    \vspace{-15pt}
    \caption{\textbf{Visualizations of Generated Refocus Videos.}
Pretrained VDMs (without finetuning) can  generate diverse blur-to-sharp transitions. These foreground-background refocusing effects are common in daily videos and films used for training, enabling VDMs to learn such dynamics to restore details progressively.}
    \label{fig:vis_explan}
    \vspace{-15pt}
\end{figure*}

To achieve these targets, we propose a novel paradigm for ultra-low-bitrate image compression that explores the dynamics in generative image decoding. Our framework reinterprets single-image generative decoding as a virtual temporal process, where a compact anchor evolves to a high-fidelity reconstructed image. As illustrated in Fig.~\ref{fig:vis_explan}, pretrained VDMs can naturally refine blurry frames into sharper frames with richer details. We exploit this blur-to-sharp temporal prior by treating the anchor as a coarse initial frame and the target image as its sharp, detail-resolved final frame. For efficiency, we collapse the original multi-frame evolution into a two-frame refinement process.

Concretely, we define an explicit intermediate state for ULB-IC: a visible anchor that preserves scene geometry, semantic layout, and coarse appearance while discarding high-frequency details. As shown in Fig.~\ref{fig:teaser} (b), this anchor is encoded and transmitted by a VAE-based Anchor Codec.
At the decoder, we first decode the anchor from the bitstream and treat it as the initial state. The evolution from anchor to fine-grained reconstruction is then modeled as a video-like progression using a pretrained video diffusion model. We construct a virtual two-frame video, where the anchor is the first frame and the original image is the second. This setup casts generative decoding as a next-frame prediction task conditioned on the anchor, guided by the dynamic priors of the VDM.
This yields a reconstruction consistent with the anchor and perceptually aligned with the source without extra control networks, improving both realism and  fidelity.

However, two key challenges arise when applying video diffusion models to image compression. (1) Domain Gap Between Video Generation and Image Decoding. Video diffusion models are primarily trained for gradual temporal evolution across dozens of frames, and emphasize scene transitions and object motions. 
In contrast, our goal is to achieve high-fidelity detail synthesis through a single-interval transition: a virtual two-frame ``video'' consisting solely of the compact anchor and the target image.
(2) High Computational Cost of Iterative Denoising. State-of-the-art VDMs often require 50 iterative denoising steps and start from standard Gaussian noise, causing high latency and injecting ambiguity that can harm fidelity \cite{aigc_corrdiff, chen2024faithdiff}.
Our aim is to turn multi-step video diffusion into a \emph{one-step} generative decoder that preserves both fidelity and realism.

To address these challenges, we propose a two-stage training scheme that adapts video diffusion models for efficient and faithful ultra-low-bitrate image decoding: \textbf{Stage I: Next-Frame Decoding Adaptation.} 
We jointly finetune the pretrained VDM and the anchor codec to synthesize high-fidelity details from compact anchor frames while suppressing undesirable temporal artifacts, such as object motion and scene flicker. This adaptation aligns the generative priors of the VDM with the anchor-conditioned detail synthesis process, enabling accurate and perceptually faithful reconstruction through a single-interval temporal evolution.
\textbf{Stage II: One-Step Generative Bypass.} To eliminate the latency introduced by iterative denoising, we collapse the multi-step diffusion process to one-step generation by establishing a latent bypass between anchor compression and video generation.
Specifically, we condition the encoding of the anchor Compression-VAE on generative latents obtained from the Video-VAE. We then introduce a  Bypass Refinement module, which is a learned mapping from the compression latent space to the generative latent space. This bypass regularizes and aligns the latent spaces of the two VAEs, reducing generative uncertainty and enabling faithful one-step decoding.

Our contributions can be summarized as follows.
\begin{enumerate}
    \item \textbf{A novel ULB-IC paradigm:} We reinterpret generative image decoding as a virtual temporal transition from a compact anchor into a faithful and realistic reconstruction.

    \item \textbf{Efficient generative decoding:} Our two-stage training pipeline adapts VDMs for compression-aware next-frame prediction while enabling one-step generation.
    
    \item \textbf{State-of-the-art performance:} Extensive experiments demonstrate that our model achieves superior perceptual realism and fidelity with relatively faster decoding.

\end{enumerate}

\section{Related Work}
\vspace{-3pt}
\label{sec:related}
\vspace{-3pt}
\subsection{Generative Image Compression}
\vspace{-3pt}
\noindent\textbf{VAE- and GAN-based Codecs.}
Previous VAE-based learned image compression models \cite{cheng2020learned, zhang2023neural, jiang2023mlic, jiang2023mlic++, minnen2020channel, li2025differentiableVMAF, aigc_add7, aigc_add8, aigc_add9} mainly optimize  objective metrics, such as PSNR and MS-SSIM. Agustsson \textit{et al.}\cite{aigc_gan0} and HiFiC \cite{aigc_hific} pioneered GAN-based methods for low-bitrate image compression and aim to improve realism and fidelity. MR \cite{aigc_MR} connects generative and non-generative compression. MS-ILLM \cite{aigc_MSILLM} introduces a local likelihood model to improve realism. EGIC \cite{aigc_egic} uses discriminators conditioned on segmentation for stronger adversarial training. Recent studies \cite{aigc_vqgan, aigc_vqgan1, aigc_glc} use vector quantization for extreme semantic compression. TACO \cite{aigc_taco} proposes a text-guided encoder to improve  fidelity.

\noindent\textbf{Diffusion-based Codecs.}
Recent works \cite{aigc_text_sketch, aigc_perco, aigc_perco_image, aigc_add4, aigc_add1, aigc_add2, aigc_add3, aigc_addd, aigc_DIRAC, aigc_HFD, aigc_cdc, aigc_FDM, aigc_onedc, aigc_adding1, aigc_addingdl, aigc_adding2, aigc_icme, aigc_dual1, aigc_picd, guo2025oscar, aigc_diffpc, zhang2026autoregressive, zhou2026dual, ling2026free} use generative priors from large pretrained controllable image diffusion models \cite{control1, control2, control3}. Building on DDPM \cite{DDPM} and LDM \cite{LDM}, they show better perceptual realism than conventional GAN-based methods. \cite{aigc_corrdiff} introduces a privileged decoder to correct the denoising process. DiffC \cite{aigc_diffc} achieves zero-shot compression using reverse-channel coding. ResULIC \cite{aigc_resulic} adds residual coding for semantic alignment. PICD \cite{aigc_picd} proposes a model for both screen content and natural images. 
Most models \cite{aigc_DIRAC, aigc_add2, aigc_perco, aigc_HFD} based on standard image diffusion models mainly inject conditions through channel concatenation, which may not be a principled way to exploit context.

\vspace{-5pt}
\subsection{Video Diffusion Model}
\vspace{-3pt}
Text-to-video (T2V) synthesis has progressed rapidly. Early work with large-scale Transformers, such as CogVideo \cite{hong2022cogvideo} and Phenaki \cite{villegas2022phenaki}, showed that the approach was feasible with great potential. Later advanced diffusion models \cite{video1, video2, video3, video4, video5} brought a clear jump in generation quality. With the development in the Diffusion Transformer (DiT) architecture, LinGen \cite{video6}, CogVideoX \cite{yang2024cogvideox} and OpenAI’s Sora \cite{openai_sora_2024} reached a new level for text-to-video generation. Recently, WAN \cite{wan2025wan}, Tencent’s HunyuanVideo \cite{kong2024hunyuanvideo} and OpenAI's Sora-2 \cite{openai_sora2} together mark the current frontier of modern video generation.

\vspace{-5pt}
\section{Methodology}
\vspace{-3pt}
\label{sec:Methodology}

\subsection{Preliminary: In-Context Learning in VDMs.}
We build upon CogVideoX-1.5~\cite{yang2024cogvideox}, a text-to-video diffusion model that generates frame sequences by progressively denoising latent noise conditioned on a text prompt. The architecture comprises: (1) a 3D causal Video-VAE for joint spatial-temporal compression whose encoder maps input videos to latents at 1/8 height and width; (2) a  text encoder~\cite{t5encoder} producing prompt embeddings; and (3) a DiT-based~\cite{dit} noise-prediction transformer. The transformer operates on the concatenation of text and video tokens with 3D attention and applies 3D rotary position embeddings \cite{su2024roformer_rope} to queries and keys for spatial-temporal modeling.

As a video foundation model trained on large-scale continuous visual data, CogVideoX exhibits emergent behaviors analogous to LLMs, including unified generation and in-context learning~\cite{lin2025realgeneral, chen2025unireal}. Given preceding frames $\{c_1,c_2,\ldots\}$ as context, it autoregressively predicts subsequent frames $\{y_1,y_2,\ldots\}$ by capturing long-range spatio-temporal dependencies. This next-token-style formulation supports  continued pretraining and task-specific finetuning, enabling coherent future-frame synthesis that remains consistent with the provided visual context.

\vspace{-3pt}
\subsection{Reframing Generative Image Compression}
\begin{wrapfigure}{r}{0.6\columnwidth}
\vspace{-25pt}
\centering
\includegraphics[width=\linewidth]{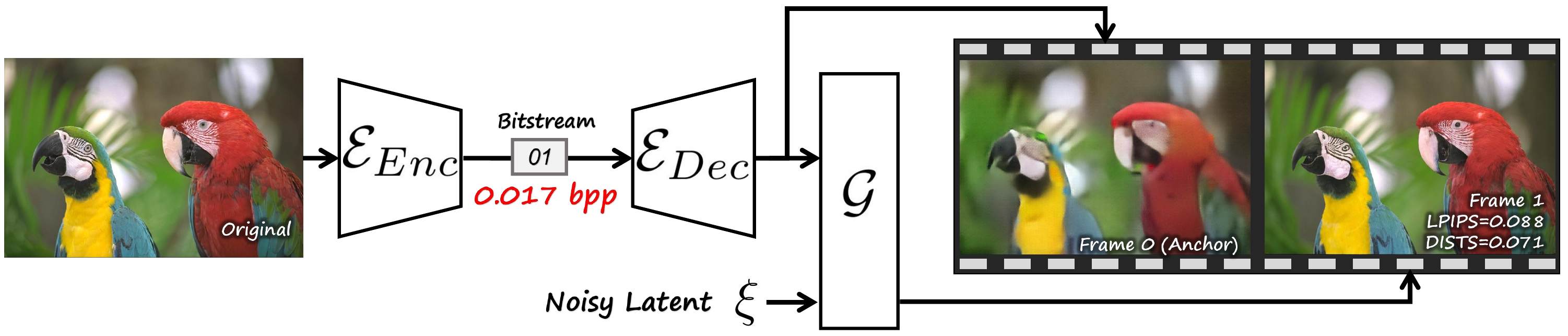}
\vspace{-21pt}
\caption{An illustration of our next-frame prediction paradigm for generative compression.}
\label{fig:illustration}
\vspace{-20pt}
\end{wrapfigure}
As discussed in Sec.~\ref{sec:intro}, we focus on modeling the transition from a compact anchor to a high-fidelity reconstruction with video diffusion priors. We adopt a two-frame abstraction: the anchor image serves as the first frame for video generation, and the target image as the second. Accordingly, we reinterpret the anchor as the conditional frame and task the VDM to predict the next frame that restores realistic details.
As shown in Fig. \ref{fig:illustration}, the encoding and generative decoding process for image $x$ can be defined as
\begin{equation}
\hat{x} = {(\mathcal{G}\circ \mathcal{E}_{Dec})}(l;\,\xi),\quad l = Q\!\big(\mathcal{E}_{Enc}(x)\big), 
\end{equation}
where $\mathcal{E}_{Enc}$ is the encoder of the Anchor Codec and $Q$ denotes quantization.
$\xi$ represents the noisy latent input of the target frame and $l$ refers to the transmitted latent. $\mathcal{E}_{Dec}$ denotes the anchor decoder, and $\mathcal{G}$ is a VDM adapted for single-interval photorealistic next-frame prediction.

In the following sections, we introduce our two-stage training scheme designed to bridge the domain gap between video generation and image decoding, and to mitigate the computational overhead of multi-step diffusion sampling.

\subsection{Why is a VDM preferred over an image diffusion model (IDM)?}
\noindent\textbf{\textit{1. Temporal Prior: VDM’s built-in transition capability.}}
\begin{wrapfigure}{r}{0.69\columnwidth}
\vspace{-20pt}
\centering
\includegraphics[width=\linewidth]{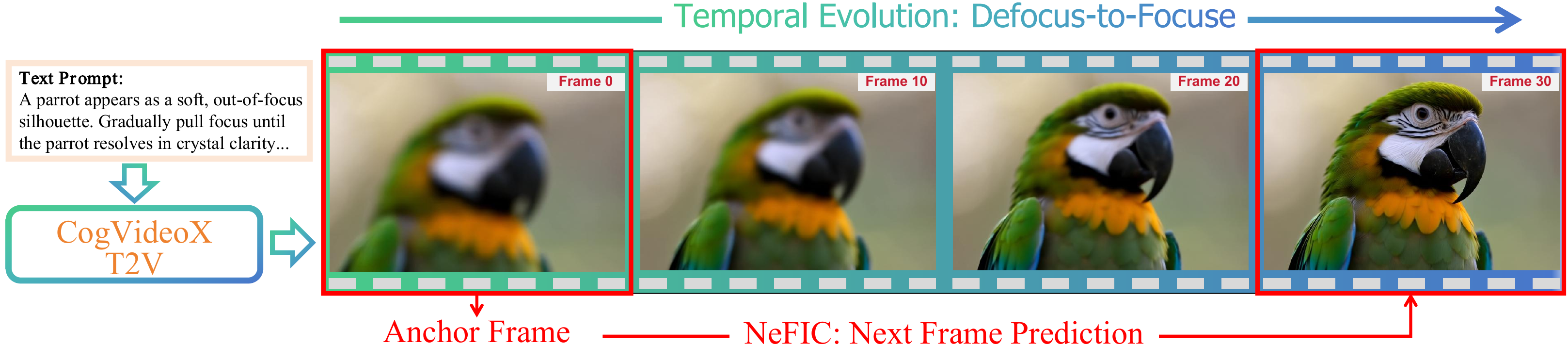}
\vspace{-19pt}
\caption{\textbf{Defocus-to-Focus Temporal Transition Prior.}
These generated frames show that VDM generates natural textures where appropriate (e.g., parrot feathers). NeFIC exploits this property and collapses multi-frame transitions into next-frame prediction.}
\label{fig:temporal_transition}
\vspace{-19pt}
\end{wrapfigure}
Our key insight is that pretrained VDMs can naturally generate \textbf{temporal defocus-to-focus  transitions}: starting from vague frames, VDMs generate high-frequency details in the subsequent frames. As shown in Fig.~\ref{fig:temporal_transition}, details of parrot feathers are temporally refined.

This behavior is not incidental. Trained on large-scale video data, VDMs internalize world knowledge and inter-object relations. 
Since video corpora contain abundant ``defocus-to-focus'' temporal patterns, VDMs learn to temporally refine coarse structure into detail-resolved imagery.

We repurpose this generative prior: heavily compressed anchors represent the early blurry frames, while the final reconstructions correspond to the clear frames at the end.
This is significantly different from predicting residuals \cite{aigc_DIRAC, aigc_add2}. \textbf{NeFIC reinterprets the blur-to-clarity temporal transition (rather than motion) in VDM as generative decoding. This pretrained transition prior is not internalized by IDMs.}

\noindent\textbf{\textit{2. Architectural Advantages: Leveraging VDM’s 3D spatiotemporal structure and in-context learning capability.}}

Inspired by in-context learning in LLMs, we reformulate VDM-based generative decoding as conditional next-frame prediction.
We fully exploit the pretrained \textbf{3D Attention} and \textbf{3D RoPE}: NeFIC performs frame/token-level concatenation to treat the anchor frame as a preceding context.

IDM-based codecs typically inject conditions by concatenating the conditional latent with the noisy latent along the channel dimension \cite{aigc_DIRAC, aigc_add2, aigc_perco, aigc_HFD}. 
Thus, conditional interactions occur only through implicit cross-channel feature mixing.
By contrast, token concatenation coupled with 3D Attention enables explicit and adaptive retrieval: each noisy token can directly attend to and query any anchor token. 
Meanwhile, 3D RoPE imposes a correspondence prior along the temporal axis, biasing attention toward spatially aligned regions across frames.
Thus, NeFIC synthesizes high-frequency details as structurally anchored refinements of the coarse layout.
\textbf{NeFIC inherits VDM’s 3D inductive bias as a more principled condition-injection pathway. Such effective conditional generation and in-context learning capabilities are not supported by standard IDMs.}

\begin{figure*}[t]
    \centering
    \includegraphics[width=0.79\linewidth]{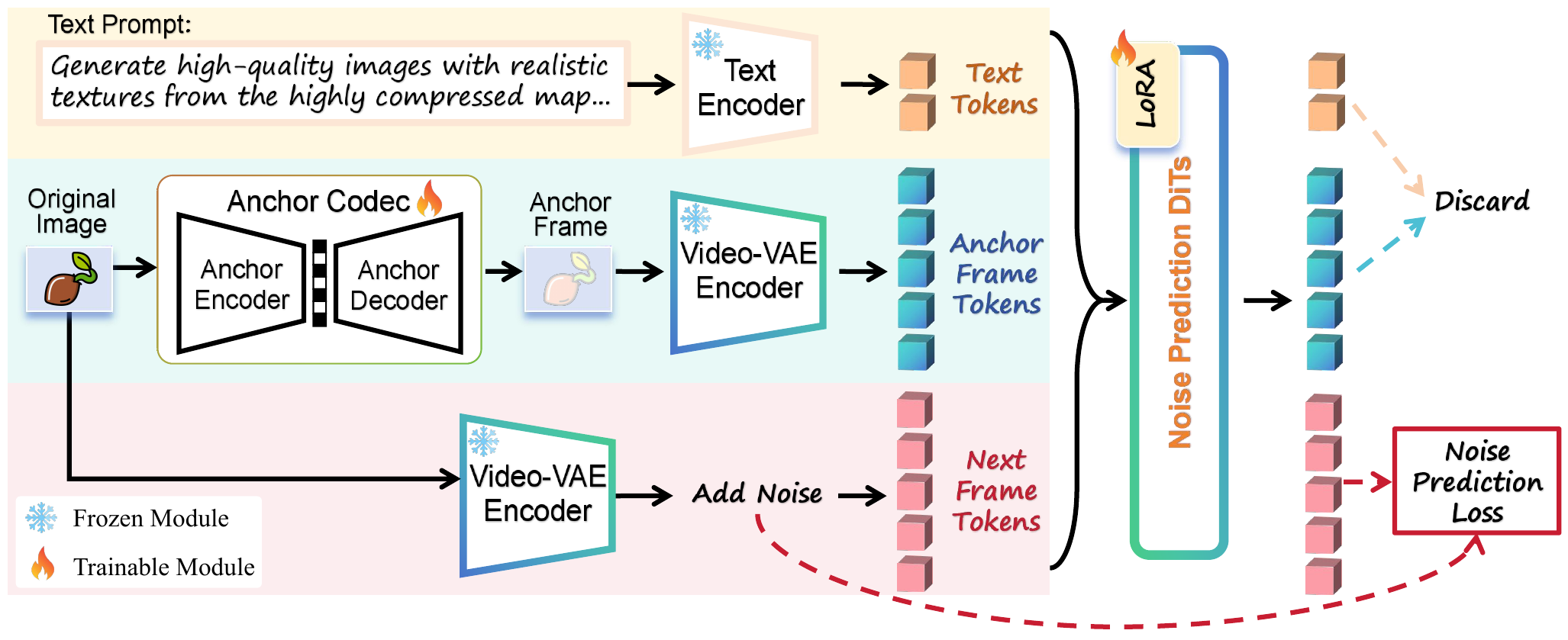}
    \vspace{-11pt}
    \caption{\textbf{Overview of Training Stage I.} The input to the DiT blocks is a concatenation of three types of tokens: (1) text embeddings, (2) tokens from the anchor frame, and (3) noised tokens from the target frame. During training, only the output corresponding to the noised target tokens is supervised using a noise prediction loss, while the other two segments are discarded.
    }
    \vspace{-15pt}
    \label{fig:pipeline1}
\end{figure*}

\vspace{-5pt}
\subsection{Stage I: Next-Frame Decoding Adaptation}
\vspace{-3pt}
We first cast generative decoding as a single-interval next-frame prediction in the latent space of a frozen Video-VAE (Fig. \ref{fig:pipeline1}). 
This stage focuses on synthesizing fine-grained details while suppressing undesirable temporal artifacts, such as object motion and scene flicker.
During training, we adapt the DiT~\cite{dit} blocks to condition on the decoded anchor, and to predict the clean latent of the target frame from its noised latent at diffusion step $t\in\{1,\dots,T\}$ with schedule $(\alpha_t,\sigma_t)$ satisfying $\alpha_t^2+\sigma_t^2=1$.
For efficiency, we finetune only the Anchor Codec and the attention projections 
$(W_q,W_k,W_v,W_{out})$ 
via LoRA~\cite{hu2022lora}, while keeping the Video-VAE and the text encoder frozen. The  video diffusion priors provide photorealistic texture and detail synthesis without auxiliary control networks.

\noindent\textbf{Tokenization and Conditioning.}
Let $E_{\text{text}}$ denote the frozen text encoder, $E_{\text{Vid}}$ the frozen Video-VAE encoder, $f_\theta$ the DiT blocks, and $\oplus$ token-sequence concatenation. 
The model consumes three sequences:
\begin{enumerate}
    \item \textbf{Text tokens:} $\mathbf{z}_{\mathrm{text}} = E_{\text{text}}(p)$, where $p$ is a universal prompt (``generate a high-quality image with realistic texture...''). Our framework eliminates the need for detailed scene descriptions, as conditions and semantics are fully transmitted via anchor frames. The  text prompt serves only as guidance of the transition behavior.
    \item \textbf{Anchor tokens:} $\mathbf{z}_{\mathrm{anchor}} = E_{\text{Vid}}\!\big(\mathcal{E}_{Dec}(l)\big)$, obtained by encoding the decoded anchor $\mathcal{E}_{Dec}(l)$, which serves as the initial-frame condition.
    \item \textbf{Noisy target tokens:} $\mathbf{z}_t$, formed by encoding the original image $x$ into $\mathbf{z}_0 = E_{\text{Vid}}(x)$ and adding noise:
    \begin{equation}
        \mathbf{z}_t = \alpha_t \mathbf{z}_0 + \sigma_t \boldsymbol{\epsilon},\quad \boldsymbol{\epsilon}\sim\mathcal{N}(\mathbf{0},\mathbf{I}).
    \end{equation}
\end{enumerate}
\noindent\textbf{Noise Prediction and Loss.}
We condition on $\mathbf{c}=\mathbf{z}_{\mathrm{text}}\oplus \mathbf{z}_{\mathrm{anchor}}$ and feed $\mathbf{c}\oplus \mathbf{z}_t$ into $f_\theta$ at step $t$. Using $v$-prediction parameterization \cite{vprediction}, we reconstruct target latent as
\begin{gather}
\mathbf{v}_\theta^{(t)}(\mathbf{z}_t,\mathbf{c})
= f_\theta(\mathbf{z}_t,\mathbf{c},t), \label{eq:v_theta_t}\\
\hat{\mathbf{z}}_{0}
= \alpha_t \mathbf{z}_{t}
- \sigma_t\, \mathbf{v}_\theta^{(t)}(\mathbf{z}_t,\mathbf{c}) \label{eq:z0_t}
\end{gather}
The output tokens corresponding to the text and anchor segments are discarded. Only the output token segment aligned with the target tokens is supervised as
\begin{equation}
    \mathcal{L}_{noise} \;=\; \left\|\hat{\mathbf{z}}_0 - \mathbf{z}_0\right\|_2^2.
\end{equation}

\noindent\textbf{Auxiliary Anchor Loss.}
To keep the decoded anchor $x_{\mathrm{anchor}}=\mathcal{E}_{Dec}(l)$ within the training distribution of the Video-VAE and maintain its conditioning strength, we add an auxiliary reconstruction term on $x_{\mathrm{anchor}}$:
\begin{equation}
\begin{split}
    \mathcal{L}_{\mathrm{aux}}
    &= \lambda_{\mathrm{MSE}} \,\big\|x - x_{\mathrm{anchor}}\big\|_2^2 \\
    &\quad + \lambda_{\mathrm{LPIPS}} \,\mathrm{LPIPS}\!\big(x,\,x_{\mathrm{anchor}}\big),
\end{split}
\end{equation}
with $\lambda_{\mathrm{MSE}}=5$ and $\lambda_{\mathrm{LPIPS}}=1$. 
The MSE term enforces pixel-level fidelity so the anchor remains stably encodable by $E_{\text{Vid}}$; the LPIPS term~\cite{lpips} preserves perceptual similarity and high-level semantics, ensuring sufficient and faithful  semantic information for conditioning.

\noindent\textbf{Overall Objective.}
The Stage~I objective combines noise prediction, anchor regularization, and a bitrate penalty:
\begin{equation}
\mathcal{L}_{\mathrm{stageI}}
= \mathcal{L}_{\mathrm{noise}} + \lambda_{\mathrm{aux}}\mathcal{L}_{\mathrm{aux}} + \lambda_R\, R,
\end{equation}
where $R$ is the bitrate loss of compressing the anchor frame and $\lambda_R$ trades rate against reconstruction quality.

This stage encourages VDM LoRAs to learn the conditional dependence in the compact anchor $\mathbf{z}_{\mathrm{anchor}}$, thereby modeling the next-frame transition and recovering a fine-grained latent $\hat{\mathbf{z}}_0$ under multi-step diffusion.

\begin{figure*}[t]
    \centering
    \includegraphics[width=1\linewidth]{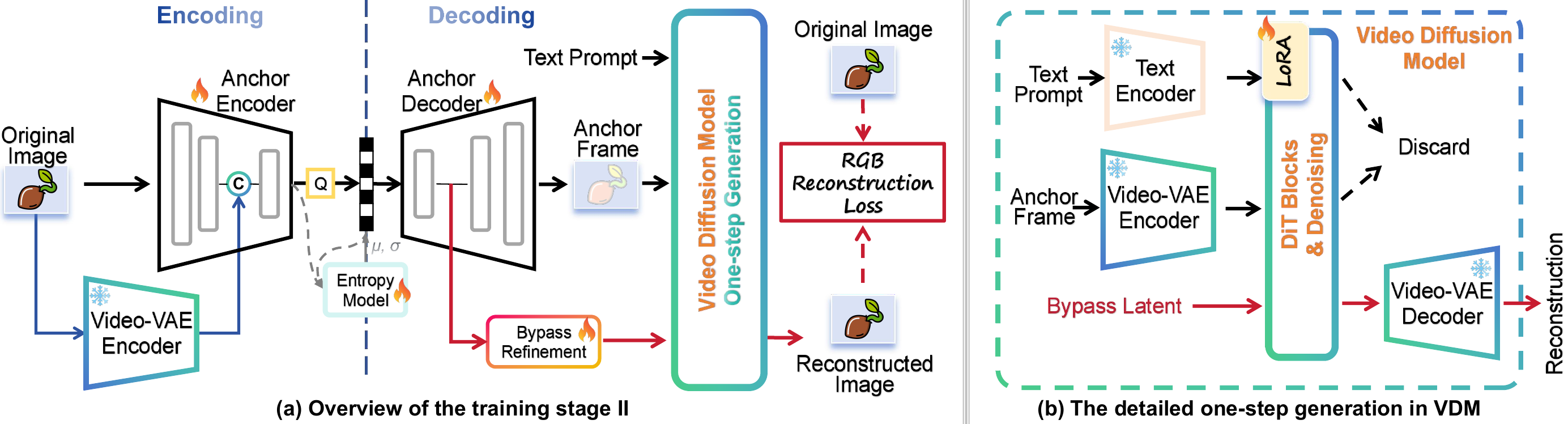}
    \vspace{-21pt}
        \caption{\textbf{Overview of Training Stage II.} \textbf{(a) Compression pipeline.} The Anchor Encoder is conditioned on Video-VAE latents of the original image and produces entropy-coded anchor latents. The Anchor Decoder reconstructs the anchor frame and a bypass latent. \textbf{(b) One-step denoising.} A single-step video diffusion model consumes text embeddings, anchor tokens, and the bypass latent to predict the target latent, which the Video-VAE decoder converts to final reconstruction.}
    \vspace{-10pt}
    \label{fig:pipeline2}
\end{figure*}

\vspace{-10pt}
\subsection{Stage II: One-Step Generative Bypass}
\vspace{-5pt}
To collapse multi-step denoising into a single step, we  initialize the target latent with an informative latent instead of Gaussian noise, which helps to reduce ambiguity and semantic drift \cite{chen2024faithdiff}. Concretely, we introduce a latent bypass that couples the Anchor Codec with VDM  through the two strategies detailed below.

\begin{wrapfigure}{r}{0.64\textwidth}
    \vspace{-25pt}
    \centering
    \includegraphics[width=\linewidth]{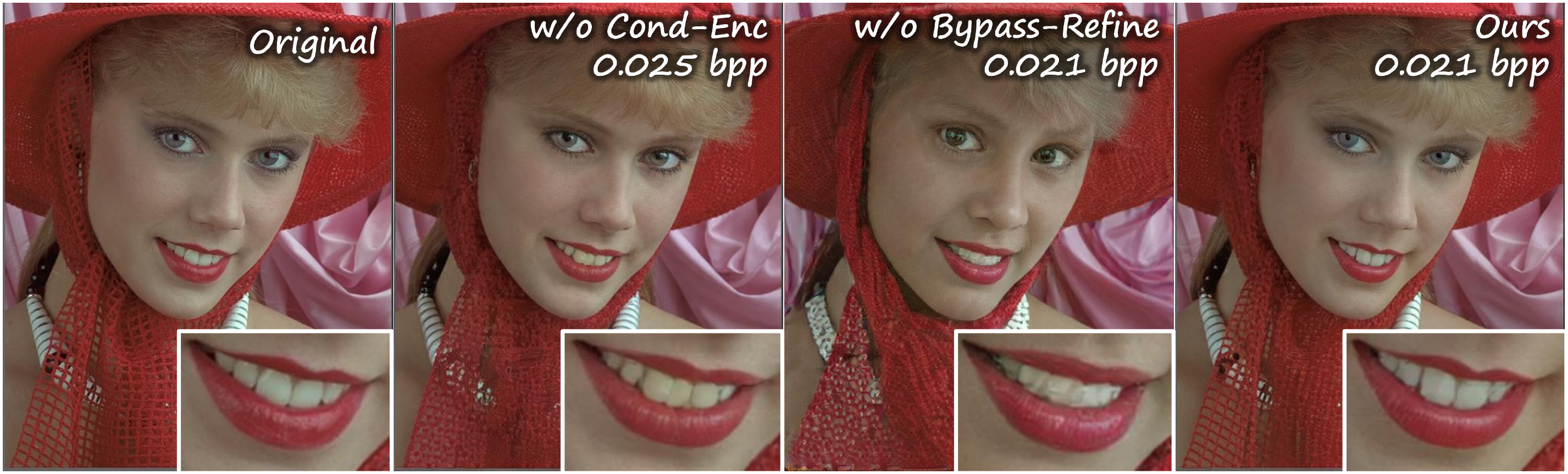}
    \vspace{-20pt}
    \caption{Reconstruction visualizations of three variants.}
    \vspace{-21pt}
    \label{fig:midabl}
\end{wrapfigure}

\noindent\textbf{Conditional Anchor Encoding.}
Due to the latent-space gap between the compression-VAE and generative-VAE~\cite{aigc_stablecodec}, raw features from the anchor codec are suboptimal for diffusion. Inspired by conditional coding~\cite{aaai_reference, li2022hybrid, li2023neural}, we condition and align the anchor encoder with the Video-VAE. The frozen $E_{\text{Vid}}$ encodes the original image $x$ to  latent
$
\mathbf{z}_0 = E_{\text{Vid}}(x).
$
As shown in Fig.~\ref{fig:pipeline2}, at the same spatial scale the Anchor encoder produces a feature map $\mathbf{h}_{\mathrm{enc}}$. We concatenate them along channels
$
\mathbf{h}_{\mathrm{cond}} = \big[\mathbf{h}_{\mathrm{enc}};\,\mathbf{z}_0\big],
$
then apply the codec transforms and entropy modeling to $\mathbf{h}_{\mathrm{cond}}$. This biases the Anchor Codec to preserve more scene information and layout semantics consistent with the diffusion prior.
As shown in Fig. \ref{fig:midabl}, removing the conditional encoding for the anchor (``w/o Cond-Enc'') leads to semantic drift, such as the appearance of extra teeth in the mouth.

\noindent\textbf{Bypass Refinement.}
We decode a bypass latent from the anchor latent to replace the pure noise at inference. Let $\mathbf{h}_{\mathrm{dec}}$ be the decoded feature from $\mathcal{E}_{Dec}$. A Transformer-based Semantic Bypass Refinement $T$ maps
$
\mathbf{z}_{\mathrm{bypass}} = T(\mathbf{h}_{\mathrm{dec}}),
$
which serves as the initial noisy latent for one-step generation, carrying semantics aligned to the generation manifold.
As shown in Fig. \ref{fig:midabl}, the absence of bypass refinement causes semantic information to become ambiguous and the reconstruction of teeth to appear vague. This occurs because the gap between the compression and generation latent spaces exacerbates the denoising difficulty for DiTs.

\noindent\textbf{One-step generation.}
As in Stage~I, we form $\mathbf{c}=\mathbf{z}_{\mathrm{text}}\oplus \mathbf{z}_{\mathrm{anchor}}$, where $\mathbf{z}_{\mathrm{text}}=E_{\text{text}}(p)$ and $\mathbf{z}_{\mathrm{anchor}}=E_{\text{Vid}}(x_{\mathrm{anchor}})$. 
To enable one-step generation, we model the bypass latent as a partially denoised state and set the denoising step $ t^\star $ to half of the maximum step value, specifically 500.
Using $v$-prediction, the DiT predicts noise and reconstructs the clean latent in a single step
\begin{gather}
\mathbf{v}_\theta(\mathbf{z}_{\mathrm{bypass}},\mathbf{c},t^\star)
= f_\theta(\mathbf{z}_{\mathrm{bypass}},\mathbf{c},t^\star), \label{eq:vtheta_bypass}\\
\hat{\mathbf{z}}_{0}
= \alpha_{t^\star}\mathbf{z}_{\mathrm{bypass}}
- \sigma_{t^\star}\,\mathbf{v}_\theta(\mathbf{z}_{\mathrm{bypass}},\mathbf{c},t^\star), \label{eq:z0_bypass}
\end{gather}
where $(\alpha_{t^\star},\sigma_{t^\star})$ is the fixed one-step schedule. The frozen Video-VAE decoder yields reconstructed image $\hat{x}=D_{\text{Vid}}(\hat{\mathbf{z}}_{0})$.

\begin{figure*}[t]
    \centering
    \includegraphics[width=1\linewidth]{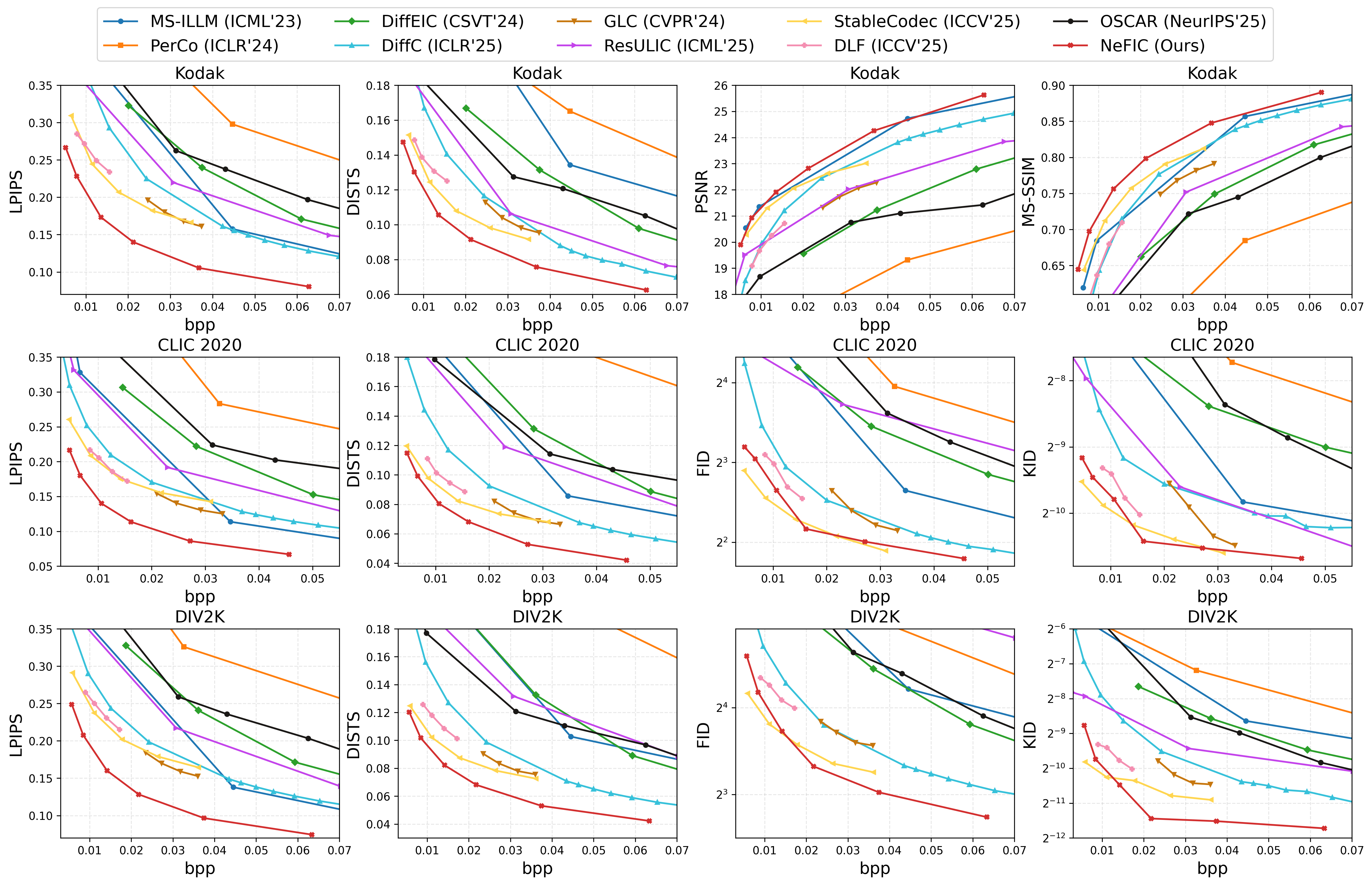}
    \vspace{-21pt}
    \caption{Rate-distortion and rate-perception curves of recent advanced methods.}
    \vspace{-15pt}
    \label{fig:curves}
\end{figure*}

\noindent\textbf{CLIP-based GAN loss.}
To leverage semantic priors from visual foundation models, we adopt a CLIP-based~\cite{clip} discriminator for adversarial training. Following~\cite{vfm_aid_gan_loss}, we train a lightweight classifier head on top while keeping the CLIP backbone frozen. The adversarial loss $\mathcal{L}_{\mathrm{GAN}}$ helps narrow the distribution gap between $x$ and $\hat{x}$~\cite{aigc_stablecodec,aigc_hific}.

\noindent\textbf{RGB reconstruction loss.}
The reconstruction loss combines the GAN term with pixelwise and perceptual criteria:
\begin{equation}
\begin{split}
\mathcal{L}_{\mathrm{RGB}}
&= \lambda_{\mathrm{GAN}}\,\mathcal{L}_{GAN}
+ \lambda_{\mathrm{MSE}}\,\|\hat{x}-x\|_2^2 \\
&\quad + \lambda_{\mathrm{LPIPS}}\,\mathrm{LPIPS}\!\big(\hat{x},\,x\big),
\end{split}
\end{equation}
with $\lambda_{\mathrm{GAN}}=1$, $\lambda_{\mathrm{MSE}}=2.5$, $\lambda_{\mathrm{LPIPS}}=0.5$.

\noindent\textbf{Overall objective.}
We retain the auxiliary anchor loss $\mathcal{L}_{\mathrm{aux}}$ and the bitrate loss $R$ from entropy model as Stage~I. The Stage~II overall training objective is 
\begin{equation}
\mathcal{L}_{\mathrm{stageII}}
= \mathcal{L}_{\mathrm{RGB}} + \lambda_{\mathrm{aux}}\,\mathcal{L}_{\mathrm{aux}} + \lambda_{R}\,R,
\end{equation}

In Stage~II, we update parameters of bypass refinement module $T$, the Anchor Codec (i.e., $\mathcal{E}_{Enc},\mathcal{E}_{Dec}$), and the LoRA adapters of $f_\theta$. Other parts remain frozen.

\section{Experiments}
\vspace{-3pt}
\label{sec:Experiments}

\subsection{Model and Training.}
\noindent\textbf{Experimental Settings.}
We use CogVideoX-1.5 as the video diffusion backbone for our model and pretrain a VAE-based Anchor Codec. Details are in the supplementary material. Our models are trained on  Flickr2W \cite{liu2020unified} with random square crops whose side length is uniformly sampled from 512 to 768 pixels. We set the batch size to 8 and train for 10k steps for stage I and 50k steps for stage II. In stage I, the learning rate is 1e-4. In stage II, we introduce the CLIP-based GAN loss at 35k steps and decay the learning rate to 1e-5 at 45k steps. The entire 2-stage training takes about 3 days. All models are optimized with AdamW.
$\lambda_{\mathrm{aux}}$ is fixed at 0.1, $\lambda_{\mathrm{R}}$ is chosen from \{5, 3, 1.7, 1.0, 0.5, 0.25\}.

\noindent\textbf{Evaluation.}
We evaluate all the methods on Kodak \cite{kodak1993}, the CLIC2020 test set \cite{CLIC2020}, and DIV2K \cite{div2k}. We report objective distortion (PSNR, MS-SSIM), perceptual fidelity metrics (LPIPS \cite{lpips}, DISTS\cite{dists}), and perceptual realism metrics (FID \cite{FID}, KID \cite{kid}). Bit-rate efficiency is measured using Bjøntegaard-Delta rate (BD-rate) \cite{bjontegaard2001}. We compute PSNR/MS-SSIM/LPIPS/DISTS on full-resolution images. Following HiFiC \cite{aigc_hific}, FID and KID are computed on 256$\times$256 patches for CLIC2020 and DIV2K. We omit FID/KID on Kodak because its small size (24 images) precludes reliable distribution estimation.

\begin{table*}[t]
  \centering
  \begingroup
  \fontsize{6pt}{8.5pt}\selectfont 
  \setlength{\tabcolsep}{1.6pt}
  \renewcommand{\arraystretch}{1.0}
  \caption{Detailed BD-Rate of different ultra-low-bitrate image compression models on Kodak, DIV2K and CLIC2020 test set.}
  \vspace{-10pt}

  \begin{tabular}{lcccccccccccc}
    \toprule
    \multirow{2}{*}{Model} &
    \multicolumn{4}{c}{Kodak} &
    \multicolumn{4}{c}{DIV2K} &
    \multicolumn{4}{c}{CLIC2020 Test} \\
    \cmidrule(lr){2-5}\cmidrule(lr){6-9}\cmidrule(lr){10-13}
     & LPIPS & DISTS & PSNR & MS-SSIM 
     & LPIPS & DISTS & FID & KID
     & LPIPS & DISTS & FID & KID \\
    \midrule
    MS-ILLM  \cite{aigc_MSILLM}        & 10.76 & 203.21 & -32.27 & -19.59 & 35.40 & 104.69 & 143.45 & 173.20 & 28.71 & 75.29 & 94.68 & 747.20 \\
    PerCo  \cite{aigc_perco}         & 112.61 & 202.54 & 303.82 & 146.77 & 266.60 & 409.65 & 176.61 & 170.99 & 326.11 & 452.83 & 128.88 & 54.05 \\
    DiffEIC  \cite{aigc_diffeic}         & 61.15 & 84.54 & 121.28 & 77.12 & 106.71 & 134.88 & 162.47 & 141.98 & 131.51 & 158.89 & 73.68 & 38.45 \\
    GLC  \cite{aigc_glc}          & -16.86 & -6.07 & 55.25 & 27.76 & -17.79 & -15.28 & 3.25 & -21.90 & -18.12 & -18.72 & -32.22 & -36.10 \\
    DiffC   \cite{aigc_diffc}          & 0.00 & 0.00 & 0.00 & 0.00 & 0.00 & 0.00 & 0.00 & 0.00 & 0.00 & 0.00 & 0.00 & 0.00 \\
    ResULIC  \cite{aigc_resulic}          & -9.28 & -6.83 & 7.60 & 24.38 & 38.02 & 68.73 & 269.67 & 0.57 & 42.08 & 64.07 & 32.80 & -62.06 \\
    StableC \cite{aigc_stablecodec}  & -42.19 & -41.20 & -22.05 & -20.82 & -27.31 & -47.56 & -52.06 & -70.08 & -27.39 & -48.35 & -70.67 & -73.46 \\
    DLF  \cite{aigc_addingdl}         & -42.31 & -36.97 & 7.77 & -1.25 & -20.07 & -33.97 & -25.71 & -34.37 & -22.51 & -36.44 & -60.82 & -67.93 \\
    
    OneDC ~\cite{aigc_onedc}          & -44.31 & -28.60 & 12.47 & 0.37 & -48.51 & -36.90 & -36.12 & - & -54.55 & -46.49 & -43.06 & - \\
    OSCAR   \cite{guo2025oscar}  & 35.16 & 28.74 & 95.69 &  51.56 & 143.25 & 43.65 & 171.66 & - & 196.58 & 97.54 & 352.81 & - \\
    \midrule
    NeFIC (Ours)        & -64.03 & -50.84 & -34.26 & -33.88 & -61.71 & -56.94 & -49.94 & -71.20 & -64.09 & -57.96 & -65.76 & -75.94 \\
    \bottomrule
  \end{tabular}
  \vspace{-13pt}
\label{tab:curves}
  \endgroup
\end{table*}

\noindent\textbf{Baselines.}
We take generative image codecs as baselines, including the VAE- and GAN-based generative codec MS-ILLM\cite{aigc_MSILLM}, GLC\cite{aigc_glc}, DLF\cite{aigc_addingdl} and recent diffusion-based methods: PerCo\cite{aigc_perco}, DiffEIC\cite{aigc_diffeic}, DiffC\cite{aigc_diffc},  ResULIC\cite{aigc_resulic}, OSCAR\cite{guo2025oscar} and StableCodec \cite{aigc_stablecodec}.

\subsection{Main Results}
\vspace{-3pt}
\subsubsection{Quantitative Evaluation}
\vspace{-3pt}
\noindent\textbf{Perceptual Metrics.}
In Fig.~\ref{fig:curves}, we plot rate-perception curves across various datasets. The corresponding BD-rate results are summarized in Tab.~\ref{tab:curves} (lower is better; negative indicates bitrate savings), where we report BD-Rate on DIV2K with DiffC as the baseline.
Compared with the earlier diffusion-based codec PerCo~\cite{aigc_perco}, our model reduces bitrate on DIV2K by \mbox{$89.82\%$}, \mbox{$91.61\%$}, \mbox{$87.94\%$}, and \mbox{$93.25\%$} on LPIPS, DISTS, FID, and KID, respectively.
Even against recent SOTA method DLF~\cite{aigc_addingdl} (ICCV'2025), our method still achieves \mbox{$51.92\%$}, \mbox{$48.16\%$}, \mbox{$31.82\%$}, and \mbox{$33.47\%$} BD-rate savings on the same four metrics.
Across datasets and metrics, our approach consistently attains substantial bitrate reductions at matched perceptual quality, indicating that it preserves both perceptual realism and image fidelity.

\begin{figure*}[t]
    \centering
    \includegraphics[width=1\linewidth]{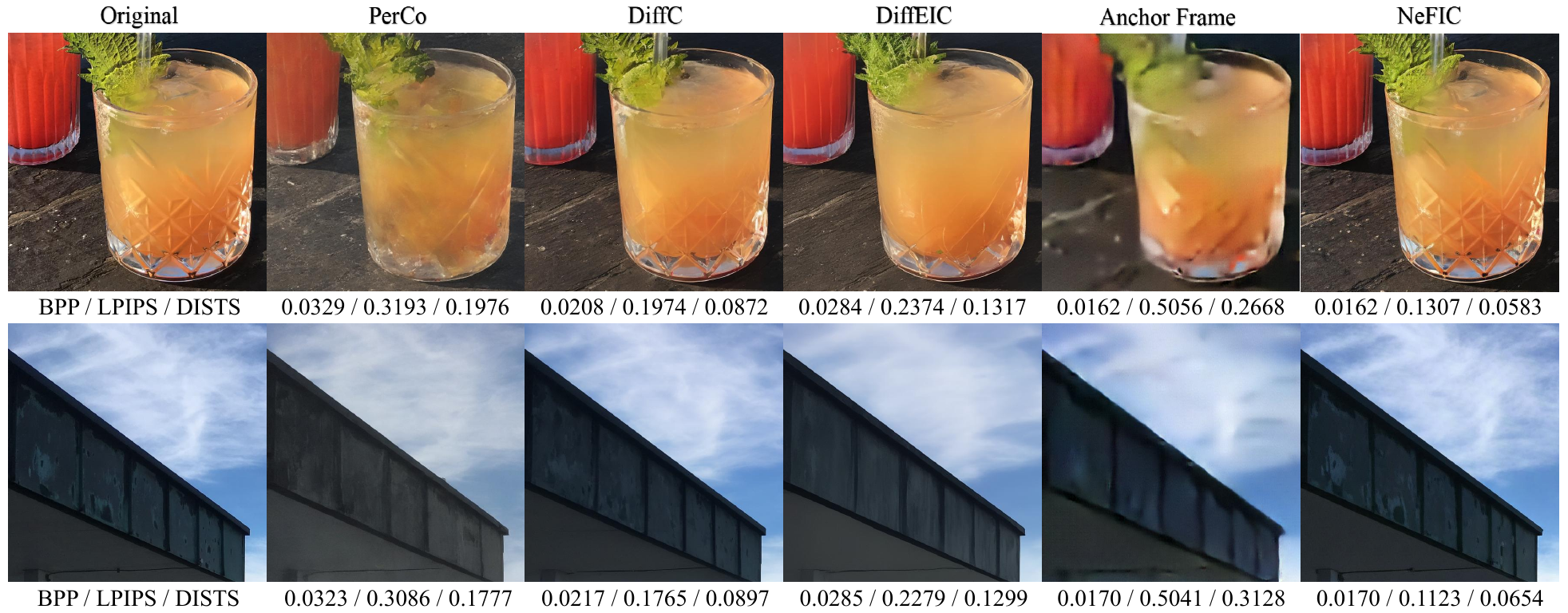}
    \vspace{-16pt}
    \caption{Qualitative examples on the CLIC2020 dataset. Zoom in for better view. More comparisons are in supplementary.}
    \vspace{-13pt}
    \label{fig:visuals}
\end{figure*}

\noindent\textbf{Distortion Metrics.}
Tab.~\ref{tab:curves} also reports BD-rate comparisons on PSNR and MS-SSIM for Kodak; the corresponding RD curves are shown in Fig.~\ref{fig:curves}.
Whereas many generative compression methods sacrifice distortion-oriented objectives for stronger generative ability, our method improves both perceptual and distortion metrics.
On Kodak, we outperform recent SOTA model StableCodec~\cite{aigc_stablecodec} (ICCV'2025) by \mbox{$18.27\%$} BD-rate on PSNR and \mbox{$22.95\%$} on MS-SSIM.
Compared to GAN-based method GLC \cite{aigc_glc} (CVPR'24), our model also can achieve 57.08\% and 46.91\% bit savings for PSNR and MS-SSIM.
These advantages in objective metrics attest to the faithfulness of our reconstructions.
Detailed PSNR and MS-SSIM results for DIV2K and CLIC2020 are provided in the supplementary.

The consistent gains in realism, fidelity, and distortion metrics stem from our evolutionary compression pipeline. The anchor frame preserves coarse yet faithful structure and key semantics, while the video diffusion prior, with its strong temporal consistency, refines them into realistic fine textures.
This yields a tight latent-pixel dual-domain coupling: the bypass latent provides a good initialization in latent space, and the anchor frame offers a pixel-domain reference that fully exploits the pretrained video consistency.

\vspace{-5pt}
\subsubsection{Qualitative Evaluation.}
We present qualitative comparisons of our NeFIC, our anchor frame, and previous diffusion-based ULB-IC models in Fig. \ref{fig:visuals}. At ultra-low bit rates, NeFIC produces details that are more visually consistent with and faithful to the original image. For instance, our model  accurately reconstructs the regular ridges on  glassware, as well as irregular corrosion patterns on walls.  Other models fail to recover those details accurately. Despite using higher bitrates than our approach, these competing models perform worse for  LPIPS, DISTS, and visual fidelity. As we described, our anchor frame preserves only the scene geometry, semantic layout, and coarse appearance, while providing a powerful temporal initialization for next-frame decoding.

\vspace{-5pt}
\subsection{The Anchor and Bypass}
\vspace{-3pt}
\begin{wrapfigure}{r}{0.52\linewidth}
\vspace{-48pt}
\centering
\includegraphics[width=1\linewidth]{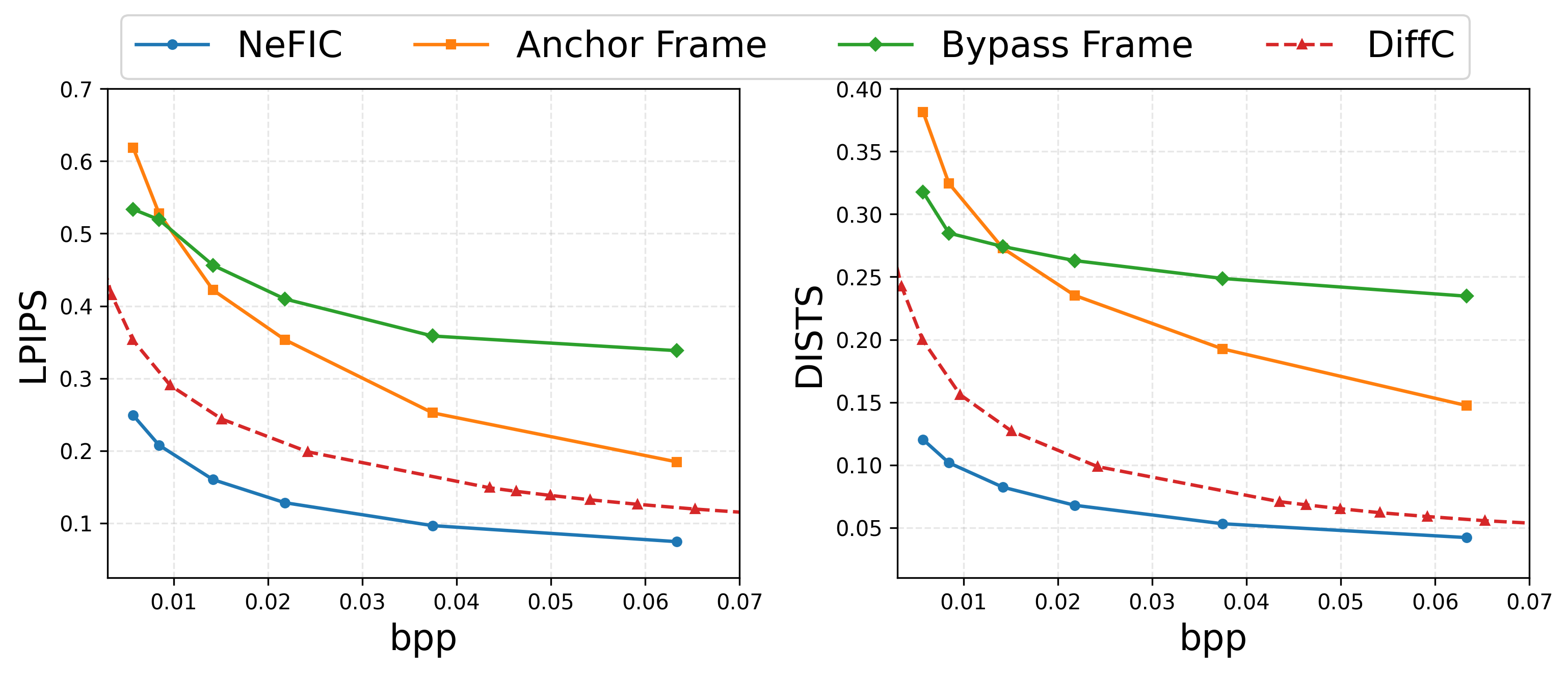}
\vspace{-22pt}
\caption{Comparisons of different frames.}
\vspace{-22pt}
\label{fig:anchor}
\end{wrapfigure}
We plot the rate-perceptual curves in Fig. \ref{fig:anchor} to compare the perceptual quality of the anchor and bypass frames (directly decoding bypass latent by Video-VAE) with the final reconstructed images. As shown, both the anchor and bypass frames exhibit lower perceptual quality, and the significant gap between their curves and the NeFIC curve highlights the effectiveness of our next-frame prediction strategy in achieving superior reconstruction fidelity with VDM.
More visualizations are provided in supplementary material.

\vspace{-8pt}
\subsection{Efficiency Evaluation.}
\vspace{-8pt}
We assess runtime efficiency as the average per-image encoding (Enc.) and decoding (Dec.) time on an A100 GPU on the Kodak and DIV2K datasets (Tab. \ref{tab:runtime_kodak_div2k}). NeFIC achieves 0.17/0.88 s (Enc./Dec.) on Kodak and 0.58/6.83 s on DIV2K, yielding 5.5-8.6× faster decoding and 1.35-4.17× faster encoding than the diffusion-based DiffEIC. 
On high-resolution DIV2K, NeFIC outperforms StableCodec by 32.5× in encoding and 6.0× in decoding due to avoiding costly tiling. 
These gains arise from single-interval prediction and one-step generation, which yield smoothly scaling latency and avoid the sharp slowdowns of multi-step diffusion or heavy tiling.
Removing the Stage II training for one-step generation increases decoding time to 30.61s for Kodak and 296.02s for DIV2K in a 50-step denoising setting. These delays are unacceptable and underscore the effectiveness and necessity of our generative bypass.

\begin{table}[t]
  \centering
  \begingroup
  \scriptsize
  \caption{Runtime (s) on Kodak and DIV2K, and BD-Rate.}
  \label{tab:runtime_kodak_div2k}
  \vspace{-7pt}
  \setlength{\tabcolsep}{2.5pt}
  \renewcommand{\arraystretch}{1.}
  \begin{tabular}{lccccc}
    \toprule
    \multirow{2}{*}{Model} &
      \multicolumn{2}{c}{Kodak (s)} &
      \multicolumn{2}{c}{DIV2K (s)} &
     \multirow{2}{*}{\shortstack{BD-Rate\\DIV2K\\(LPIPS,\%)}}\\
    \cmidrule(lr){2-3}\cmidrule(lr){4-5}
     & Enc. & Dec. & Enc. & Dec. & \\
    \midrule
    PerCo \cite{aigc_perco}    & 0.39 & 1.92 & 0.62 & 38.29 & $266.60$ \\
    DiffEIC \cite{aigc_diffeic} & 0.23 & 4.82 & 2.42 & 58.46 & $106.71$ \\
    DiffC \cite{aigc_diffc}          & 0.6-9.1 & 2.9-8.4  & 3.4-47.2 & 15.8-37.7 & $0.00$ \\
    StableCodec \cite{aigc_stablecodec}          & 0.11 & 0.14 & 18.86 & 40.63 & $-27.31$ \\
    \midrule
    w/o Stage II     & 0.16 & 30.61  & 0.51 & 296.02 & $-47.01$ \\
    \cellcolor{gray!20}\textbf{NeFIC(Ours)}      &
    \cellcolor{gray!20}{0.17} &
    \cellcolor{gray!20}{0.88}  &
    \cellcolor{gray!20}{0.58} &
    \cellcolor{gray!20}{6.83} &
    \cellcolor{gray!20}{$-61.71$} \\
    \bottomrule
  \end{tabular}
  \vspace{-17pt}
  \endgroup
\end{table}

\begin{table}[t]
  \centering
  \begingroup
  \scriptsize
  \caption{Ablation BD-rate results on different components.}
  \label{tab:ablation}
  \vspace{-7pt}
  \setlength{\tabcolsep}{3.pt}
  \renewcommand{\arraystretch}{1.}
  \begin{tabular}{lcccc}
    \toprule
    Model variants & LPIPS & DISTS & FID & KID \\
    \midrule
    w/o Interval  Adaptation         & -52.32 & -44.71 & -40.32 & -62.88 \\
    w/o One-step Bypass          & -47.01 & -42.03 & -53.72 & -77.91 \\
    \midrule
    w/o Conditional Encoding    & -56.81 & -52.34 & -42.01 & -62.46 \\
    w/o Bypass Refinement     & -48.01 & -47.19 & -39.36 & -57.91 \\
    \midrule
    w/o Aux Anchor Loss       & -19.78 & -20.46 & -31.07 & -40.74 \\
    w/o CLIP-based GAN        & -63.09 & -64.02 & -30.31 & -30.07 \\
    \cellcolor{gray!20}\textbf{NeFIC (Ours)}  &
    \cellcolor{gray!20}{-61.71} &
    \cellcolor{gray!20}{-56.94} &
    \cellcolor{gray!20}{-49.94} &
    \cellcolor{gray!20}{-71.20} \\
    \bottomrule
  \end{tabular}
  \vspace{-7pt}
  \endgroup
\end{table}

\begin{table}[t]
  \centering
  \scriptsize 
  \caption{Ablation BD-rate results on LoRA Rank and Alpha.}
  \label{tab:lora ablation}
  \vspace{-7pt}
  \setlength{\tabcolsep}{3.6pt} 
  \renewcommand{\arraystretch}{1.} 
  \begin{tabular}{lcccc}
    \toprule
    LoRA Rank/Alpha & LPIPS & DISTS & FID & KID \\
    \midrule
    64/64        & -44.18 & -41.01 & -59.91 & -82.99 \\
    128/128        & -54.82 & -51.36 & -58.34 & -80.26 \\
    256/256        & -61.71 & -56.94 & -49.94 & -71.20 \\
    \bottomrule
  \end{tabular}
  \vspace{-15pt}
\end{table}

\vspace{-8pt}
\subsection{Ablation Studies.}
\vspace{-8pt}
We ablate the key design components of NeFIC. Unless otherwise noted, all variants are evaluated on DIV2K using LPIPS, DISTS, FID, and KID. We report BD-Rate (\%, lower is better) on DIV2K with DiffC as the baseline.

\noindent\textbf{Stage I (Interval Adaptation).}
Training directly with Stage~II is possible but suboptimal. CogVideoX is pretrained on long video sequences ($>$ 25 frames) with diverse scene transitions and motions~\cite{yang2024cogvideox}. If we skip Stage~I, the LoRA must learn compression-aware single-interval prediction and single-step generation at the same time, which makes optimization harder and more unstable. In Tab.~\ref{tab:ablation}, removing Stage~I (``w/o Interval Adaptation'') degrades BD-Rate on all four metrics by 8-10\%.

\noindent\textbf{Stage II (One-step Bypass).}  
With Stage~I alone, the model supports next-frame decoding for ultra-low bitrate compression but still requires multi-step denoising. Adding Stage~II enables end-to-end training and further improves fidelity, at the cost of slightly lower realism. Compared with the Stage~I-only setting (``w/o One-step Bypass''), Stage~II improves LPIPS and DISTS BD-Rate by about 15\%, while FID and KID worsen by roughly 4\% and 7\%, indicating a mild trade-off between perceptual fidelity and realism.

\noindent\textbf{Conditional Encoding.}
The semantic latent bypass consists of \emph{Conditional Encoding} and \emph{Bypass Refinement}. Removing Conditional Encoding hampers the alignment of the bypass latent with the VDM generation space and its cooperation with the frozen Video-VAE decoding, yielding BD-rate drops about $4.9\%$ for LPIPS and $7.9\%$ for FID  (Tab.~\ref{tab:ablation}).

\noindent\textbf{Bypass Refinement.}
Ablating the bypass refinement causes a larger gap: BD-Rate degrades by ${\sim}10$-14\% on LPIPS/FID/KID. This suggests the raw VAE-decoded latent still mainly supports anchor-frame RGB reconstruction, while refinement better adapts it to the generation space with stronger semantics.

\noindent\textbf{Auxiliary Anchor Loss.}
Without the auxiliary anchor loss, the anchor frame diverges from the natural image domain, exhibiting significant semantic and color drift. This misalignment with the training distribution of Video-VAE weakens the conditioning for next-frame generation for single-interval, one-step decoding. As shown in Tab.~\ref{tab:ablation}, the absence of this loss leads to a substantial degradation in perceptual compression performance, with BD-Rate degrading to $-19.78\%$ for LPIPS and $-40.74\%$ for KID.

\noindent\textbf{CLIP-guided GAN Loss.}
Ablating the CLIP-based adversarial loss slightly improves LPIPS/DISTS but severely harms realism (FID/KID worsen by about 20 and 41 BD-Rate points). Adversarial training is therefore important for preserving generation capability and generative realism.

\noindent\textbf{LoRA Settings.}
Tab. \ref{tab:lora ablation} shows that increasing rank/alpha from $64/64$ to $256/256$ consistently improves LPIPS ($-44.18\%\!\to\! -61.71\%$) and DISTS ($-41.01 \%\!\to\! -56.94\%$) while degrading FID and KID by about 10\% in BD-rate. We adopt 256/256 as the default setting for LoRA adapters in NeFIC.

\begin{table}[t]
  \centering
  \scriptsize
  \caption{Paradigm and Foundation Model Comparison.}
  \label{tab:image comparison}
  \vspace{-7pt}
  \setlength{\tabcolsep}{3.6pt} 
  \renewcommand{\arraystretch}{1.}
  \begin{tabular}{lcccc}
    \toprule
    Generation Model & LPIPS & DISTS & FID & KID \\
    \midrule
    (1) Flux-dev-12B \cite{flux2024_dev} (w/o anchor)        & -32.01 & -33.21 & -53.97 & -72.87 \\
    
      (2) Flux-dev-12B \cite{flux2024_dev} (channel concatenation)  & -35.77 & -35.13 & -55.92 & -72.09 \\
      (3) Flux-dev-12B \cite{flux2024_dev} (token concatenation) & -42.18 & -42.31 & -53.98 & -73.19 \\
    CogVideoX-5B (Ours)       & -61.71 & -56.94 & -49.94 & -71.20 \\
    \bottomrule
  \end{tabular}
  \vspace{-15pt}
\end{table}

% \vspace{-3pt}
\noindent\textbf{Comparison with Image-Diffusion Paradigm.}
To highlight the benefit of anchor-guided next-frame decoding with a VDM, we compare against a text-to-image (T2I) diffusion codec baseline that replaces our CogVideoX-1.5-5B decoder with Flux-dev-12B~\cite{flux2024_dev}. We fine-tune this larger model under three variants: (1) decode the bypass latent in a single step without the anchor branch; (2) concatenate the anchor latent along the channel dimension; (3) concatenate anchor latent as tokens (with 2D attention only, since 3D RoPE and attention is unsupported by IDM).
As shown in Tab.~\ref{tab:image comparison}, the BD-rate results indicate that the anchor condition is not effectively exploited in variants (1)–(3). 
The T2I baseline benefits from both larger model capacity and the fact that state-of-the-art T2I models often outperform T2V models in visual realism. Our method attains substantially lower BD-rates on LPIPS and DISTS, showing higher perceptual fidelity and better structural consistency with the source image.

We attribute these gains to the explicit anchor and virtual temporal evolution, which together enforce a strong latent–pixel dual-domain coupling. Concretely, the bypass latent provides a strong initialization in latent space, while the decoded anchor frame supplies a pixel-domain reference and leverages the pretrained VDM’s learned temporal consistency to mitigate semantic drift.
These comparisons suggest that scaling parameters alone cannot achieve such impressive improvements. Additionally, what we propose is a VDM-based \textbf{paradigm}, which can benefit from more efficient future VDM backbones.

\textbf{More discussions about VDM, visualizations, and ablations are provided in the supplementary.}

\section{Conclusion}
We recast ultra-low-bitrate image compression as a virtual temporal evolution from a compact anchor frame to the final, fine-grained reconstruction. We introduce a generative next-frame decoder that leverages pretrained video diffusion priors to synthesize faithful, realistic images. To enable efficient generative decoding, we propose a two-stage training scheme: first, we adapt the VDM’s multi-frame generation to a next-frame prediction task tailored for image compression; second, we collapse the multi-step diffusion process into single-step generation. Our ULB-IC model NeFIC achieves superior perceptual compression performance with efficient decoding.

\clearpage
\newpage
% \appendix
\section*{Acknowledgment}
This work was supported by the National Key Research and Development Program of China under Grant 2024YFF0509700, the National Natural Science Foundation of China (62471290,62431015,62331014) and the Fundamental Research Funds for the Central Universities.

\bibliographystyle{splncs04}
\bibliography{main}
\end{document}